\documentclass{sig-alternate}
\usepackage{amsfonts,amssymb}
\usepackage{algorithm}
\usepackage{algorithmic}
\usepackage{url}
\usepackage{ifpdf}

\usepackage[
  pass,
]{geometry}

 \newfont{\mycrnotice}{ptmr8t at 7pt}
 \newfont{\myconfname}{ptmri8t at 7pt}
 \let\crnotice\mycrnotice%
 \let\confname\myconfname%

 \permission{Permission to make digital or hard copies of all or part of this work for personal or classroom use is granted without fee provided that copies are not made or distributed for profit or commercial advantage and that copies bear this notice and the full citation on the first page. Copyrights for components of this work owned by others than ACM must be honored. Abstracting with credit is permitted. To copy otherwise, or republish, to post on servers or to redistribute to lists, requires prior specific permission and/or a fee. Request permissions from permissions@acm.org.}
 \conferenceinfo{MM'14,}{November 3--7 2014, Orlando, FL, USA.}
 \copyrightetc{Copyright 2014 ACM \the\acmcopyr}
 \crdata{978-1-4503-3063-3/14/11\ ...\$15.00.\\
 \url{http://dx.doi.org/10.1145/2647868.2654912}
 }

\newcommand{\INPUT}{\item[\myinput]}
\newcommand{\myinput}{\textbf{Initialization:}}

\newcommand{\MYWHILE}{\item[\mywhile]}
\newcommand{\mywhile}{\textbf{repeat}}

\newcommand{\MYENDWHILE}{\item[\myendwhile]}
\newcommand{\myendwhile}{\textbf{until}}

\begin{document}
\title{3D Human Activity Recognition with Reconfigurable Convolutional Neural Networks}
\numberofauthors{1} 
%
\author{
%
%
\alignauthor
    Keze Wang$^{1}$, Xiaolong Wang$^{1}$, Liang Lin$^{1}$\thanks{Corresponding author is Liang Lin. This work 
was supported by the Hi-Tech Research and Development Program of China (no.2013AA013801), Guangdong Natural Science Foundation (no.S2013050014548), Program of Guangzhou Zhujiang Star of Science and Technology (no.2013J2200067), Special Project on Integration of Industry, Educationand Research of Guangdong (no.2012B091100148), and Fundamental Research Funds for the Central Universities.}, Meng Wang$^{2}$, Wangmeng Zuo$^{3}$\\
\affaddr
   $^1$Sun Yat-Sen University, Guangzhou 510006, China\\
   $^2$School of Computer Science and Information Engineering, Hefei University of Technology\\
   $^3$School of Computer Science and Technology, Harbin Institute of Technology\\
\email{linliang@ieee.org; \{kezewang,dragonwxl123,eric.mengwang,cswmzuo\}@gmail.com}
}

\maketitle
\begin{abstract}
Human activity understanding with 3D/depth sensors has received increasing attention in multimedia processing and interactions. This work targets on developing a novel deep model for automatic activity recognition from RGB-D videos. We represent each human activity as an ensemble of cubic-like video segments, and learn to discover the temporal structures for a category of activities, i.e. how the activities to be decomposed in terms of classification. Our model can be regarded as a structured deep architecture, as it extends the convolutional neural networks (CNNs) by incorporating structure alternatives. Specifically, we build the network consisting of 3D convolutions and max-pooling operators over the video segments, and introduce the latent variables in each convolutional layer manipulating the activation of neurons. Our model thus advances existing approaches in two aspects: (i) it acts directly on the raw inputs (grayscale-depth data) to conduct recognition instead of relying on hand-crafted features, and (ii) the model structure can be dynamically adjusted accounting for the temporal variations of human activities, i.e. the network configuration is allowed to be partially activated during inference. For model training, we propose an EM-type optimization method that iteratively (i) discovers the latent structure by determining the decomposed actions for each training example, and (ii) learns the network parameters by using the back-propagation algorithm.  Our approach is validated in challenging scenarios, and outperforms state-of-the-art methods. A large human activity database of RGB-D videos is presented in addition.
\end{abstract}
\vspace{-4mm}

\vspace{-3mm}

\vspace{-3mm}
\section{Introduction}

In the research of multimedia, there is a particular interest in the last decades on developing intelligent systems of human activity understanding with different application backgrounds, e.g. intelligent surveillance, robotics and video content search. Recently developed 3D/depth sensors have opened up new opportunities with enormous commercial values, which provide more rich information compared with the traditional cameras (e.g., three-dimensional structure information of scenes and subjects/objects). Building upon these technologies, human poses can be easily access to, and modeling complicated human activities becomes facilitated.

\begin{figure}[!htb]
\centering
\includegraphics[width=3.1in]{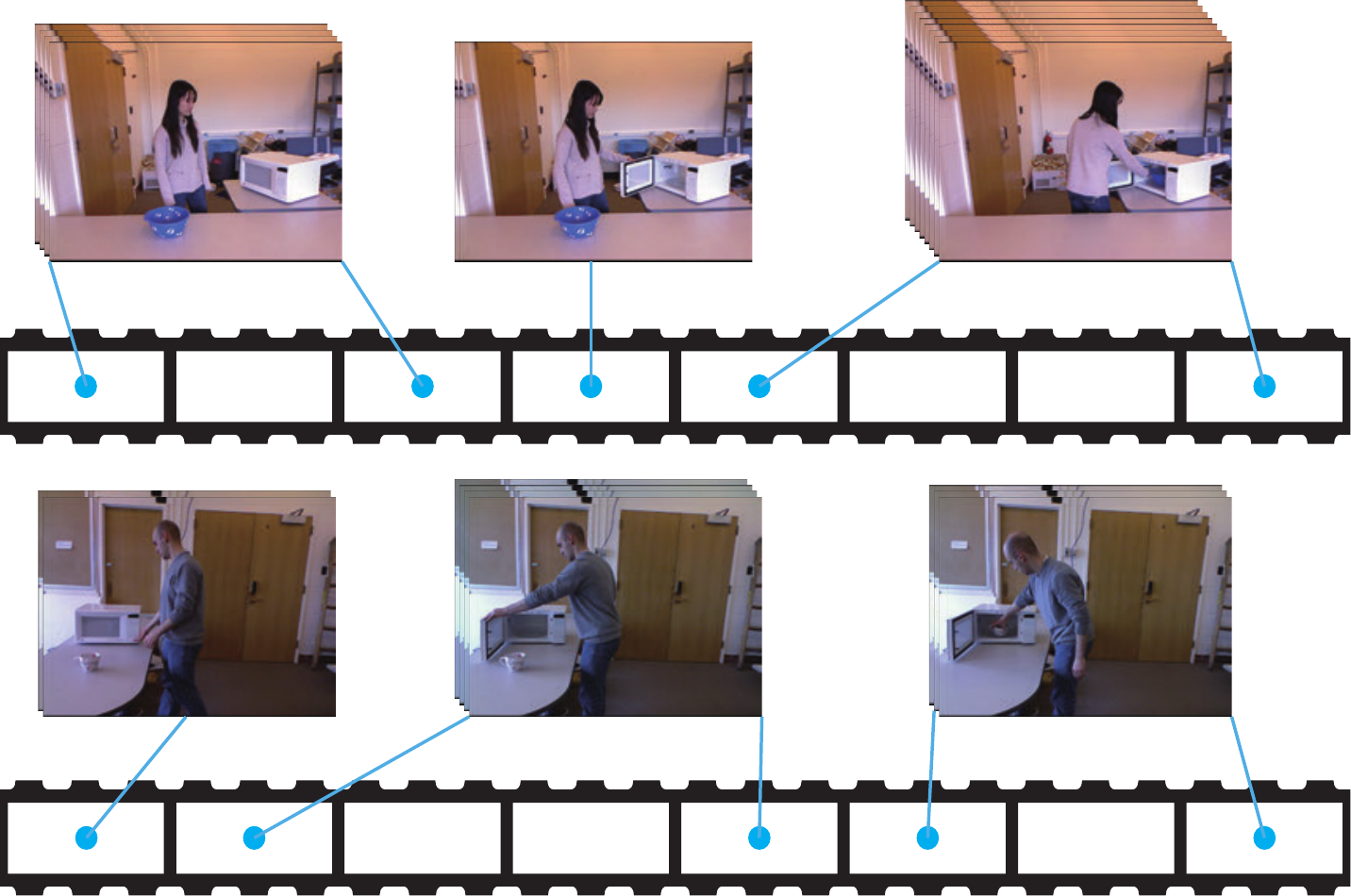}
\caption{Two activities of the same category. We consider one activity as a sequence of actions occurred over time, and temporal compositions of actions are diverse by different subjects.}\label{fig:motivation}
\end{figure}

This paper focuses on recognizing complex human activities from RGB-D videos that are captured by a Microsoft Kinect camera. There exist two main difficulties despite the additionally provided depth information:

\begin{itemize}
    \item Representing complex human appearance and motion information. Due to diverse poses/views of individuals, it is usually hard to retrieve accurate body information with motions. The depth maps are often unavoidably contaminated~\cite{HON4D}, and may become unstable due to sensor noises or self-occlusions of bodies.

    \item Capturing large temporal variations of human activities. An activity can be considered as a sequence of actions occurred over time~\cite{CIVU2013survey}. For instance, the activity of ``microwaving food'' can be decomposed into several actions such as picking up food , walking and operating microwave. The variance of a category of human activities can be hence very large, as it is uncertain how the activities to be decomposed in the temporal domain. Figure~\ref{fig:motivation} shows two activities belonging to the same category, where the temporal lengths of decomposed actions are variant by different subjects.
\end{itemize}

Most of previous methods recognize 3D human activities by training discriminative or generative classifiers based on carefully designed features~\cite{HOJ3D,HON4D,DSTIP,WuYingCVPR2012}. These approaches often require sufficient domain knowledge and heavy burden of feature engineering, which could limit their applications. Some compositional methods~\cite{WangPAMI2011,CIVU2013survey} attempt to model activities by representing videos as sequences of fixed length temporal segments. However, they may have problems on handling complex activities composed by actions of diverse temporal durations, e.g., the example in  Figure~\ref{fig:motivation}.

\begin{figure*}[!ht]
\centering
\includegraphics[width=7.0in]{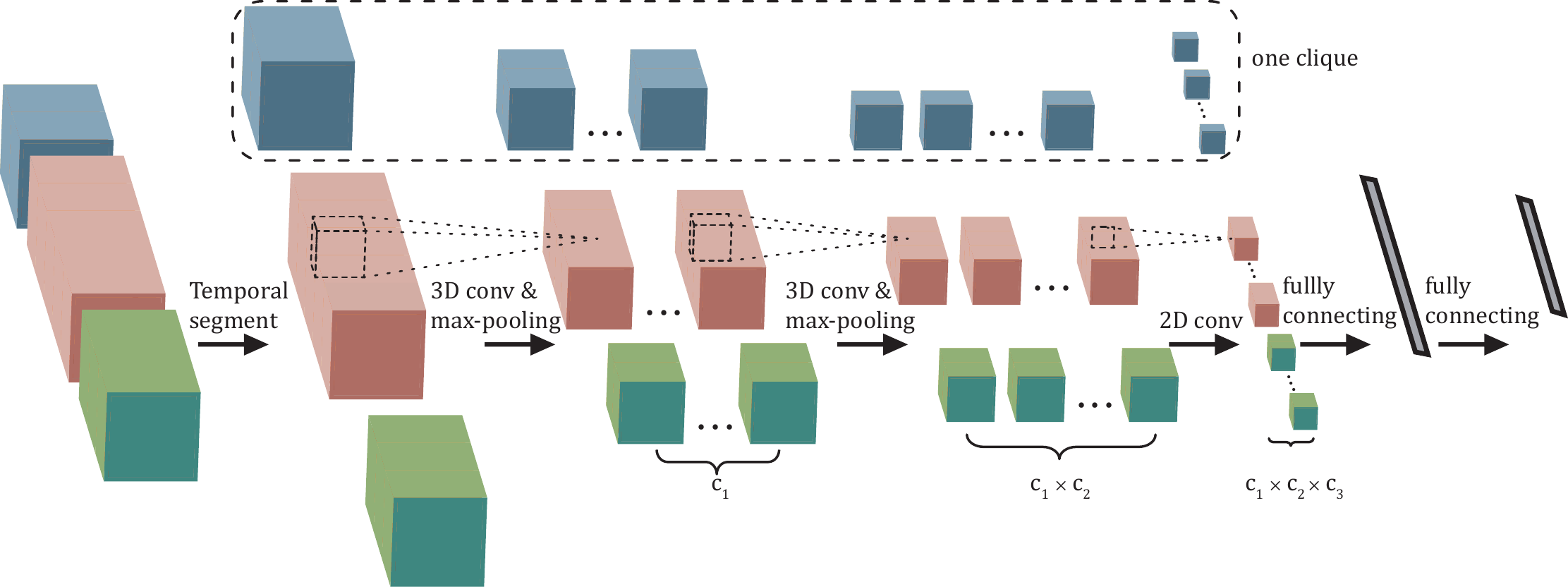}
\caption{ The architecture of our deep structured model.  The network is stacked up by convolutional layers, max-pooling operators and full connection layers, where the raw segmented videos are treated as the input. A clique is defined as a subpart of the network stacked up for several layers, extracting features for one segmented video. Moreover, the architecture can be partially enabled to explicitly handle different temporal compositions of the activities. }\label{fig:Architecture}
\end{figure*}

In this work, we develop an expressive configurable human activity model to address the above mentioned issues, absorbing the powers of two promising techniques of broad interests: deep learning~\cite{CNN1990,Hinton06,ImagenetNIPS2012,3DCNNPAMI, MDLACM13, PPDDNN13,AOGCVPR2013} and reconfigurable part-based models~\cite{AOGZhu2006, SinisaSPNCVPR2012, AOGICCV2011,LinGrammar}. We represent one human activity as a sequence of separated actions, each of which is associated with a cubic-like video segment of unfixed length, as Figure~\ref{fig:Architecture} illustrates, and learn to discover the temporal structures for a category of human activities in terms of classification. In brief, our model is built upon the deep convolutional neural networks (CNNs)~\cite{CNN1990,3DCNNPAMI}, and we allow the network to be reconfigured to capture the varying temporal compositions of activities.  We thus regard our model as a deep structured model, as it incorporates structure alternatives into a deep architecture. We consider following advantages of the deep structured model for 3D human activity recognition.

First, the deep architecture enables us to act directly on grayscale-depth data rather than relying on hand-crafted features. We build the layered network stacked up by convolutional layers, max-pooling operators and full connection layers, where the raw segmented videos are treated as inputs. We firstly apply the 3D convolutional kernel~\cite{3DCNNPAMI} over the bottom to extract features from both spatial and temporal domains, thereby encoding the motion information over adjacent frames. The 2D convolutions are then deployed upon to abstract the higher-level information. The convolutional layers for the segmented videos are computed independently to each other, in order to generate features for actions within the video segments. Afterwards, the convolution results coming from different segments are merged together into two full connection layers, giving rise to the activity classification.

Second, the structure of our model can be flexibly adjusted during inference, which is a key to improve the capability of modeling complex patterns~\cite{AOGZhu2006,AOGICCV2011,LiangMM2013}. Specifically, in each convolutional layer, we impose the latent variables to manipulate the activation of neurons, so that the network can be partially enabled to explicitly handle large temporal variations of activities. For example, some of the neurons can be turned-off to adapt the different temporal durations of separated actions. During the inference for activity recognition, we aggregate the responses in each layer of network while searching for the optimal network configuration. It is worth mentioning that we can conduct the inference in a parallel manner using GPU (Graphic Processing Unit) programming, in order to counterbalance the extra computational demand.

Training the structured deep model is another innovation of this work, as it is required to simultaneously optimize parameters and latent structure in the deep architecture.  Thus we propose an EM-type optimization method, namely Latent Structural Back Propagation (LSBP), which iterates with two steps: (i) Fixing the current model parameters, it performs activity classification while discovering the temporal composition (i.e. determining the separated actions) for each training example. (ii) Fixing the decompositions of input videos, it learns the parameters in each layer of the network using the back-propagation algorithm.

Moreover, collecting RGB-D data is relatively expensive in practice, while the amount of training data plays a critical role in deep feature learning~\cite{ImagenetNIPS2012}. Thus we propose to pre-train the network on the common RGB video data, taking advantage of existing databases of human activities. The trained parameters are then transferred into our model as the initializations.

The key contribution of this work is a novel deep structured model. To the best of our knowledge, it is original in literature to make the deep architecture reconfigurable to adaptively account for data variations. We demonstrate superior performances over state-of-the-art approaches in several challenging scenarios. In addition, we construct a new database of RGB-D data, which includes $1180$ human activities of $20$ categories.

This paper is organized as follows. Section 2 presents a review of related work. Then we present our deep structured model in Section 3, followed by a description of model learning algorithm in Section 4. Section 5 discusses the inference procedure. The experimental results and comparisons are exhibited in Section 6. Section 7 concludes this paper.


\section{Related Work}

A batch of works on human action/activity understanding mainly focused on developing robust and descriptive features~\cite{DSTIP,diMM13,HON4D,FG2013,BagSIFT,DMMM12, 3dsiftMM07}. Xia and Aggarwal~\cite{DSTIP} extracted spatio-temporal interest points from depth videos (DSTIP) and developed a depth cuboid similarity feature (DCSF) to model human activities. Oreifej and Liu~\cite{HON4D} proposed to capture spatio-temporal changes of activities by using a histogram of oriented 4D surface normals (HON4D). Most of these methods, however, overlooked detailed spatio-temporal structure information, and limited in periodic activities.

Several compositional approaches were studied for complex scenarios and achieved substantial progresses\cite{WangPAMI2011,AlanCVPR2013,MMPose2013,CVPR12PoseObject,JCorsoCVPR2012,WuYingCVPR2012,shuichengMM2011,PRLEvent}, and they decomposed an activity into deformable parts and enriched the models with contextual information. For instance, Wang et al.~\cite{WangPAMI2011} recognized human activities in common videos by training the hidden conditional random fields in a max-margin framework. For activity recognition in RGB-D data, Packer et al.~\cite{CVPR12PoseObject} employed the latent structural SVM to train the model with part-based pose trajectories and object manipulations. An ensemble model of actionlets were studied in \cite{WuYingCVPR2012} to represent 3D human activities with a new feature called local occupancy pattern (LOP).  To handle more complicated activities with large temporal variations, some powerful models~\cite{FeifeiCVPR2012,WuYingICCV2013,SinisaICCV2011} further discovered temporal structures of activities by localizing sequential actions. For example, Wang and Wu~\cite{WuYingICCV2013} proposed to solve the temporal alignment of actions by maximum margin temporal warping. Tang et al.~\cite{FeifeiCVPR2012} captured the latent temporal structures of 2D activities based on the variable-duration hidden Markov model. Koppula and Saxena~\cite{SaxenaICML2013} applied the Conditional Random Fields to model the sub-activities and affordances of the objects for 3D activity recognition.

Recently, the reconfigurable models were developed in the form of And-Or graphs~\cite{AOGZhu2006,AOGICCV2011,LiangMM2013,JVTPMM14,AOGCVPR2013}, and yielded competitive performance in several challenging scenarios. The key idea of these approaches was to discover different ways of compositions by making the models reconfigured during learning and inference. Zhu and Mumford~\cite{AOGZhu2006} first explored the And-Or graph models for image parsing. Pei et al.~\cite{AOGICCV2011} then introduced the models for video event understanding, but their approach required elaborate annotations. Liang et al.~\cite{LiangMM2013} proposed to automatically train the reconfigurable action model by a non-convex formulation. However, the above mentioned models were built on hand-crafted features.

On the other hand, the past few years have seen a resurgence of research in the design of deep neural networks, and impressive progresses were made on learning image features from raw data~\cite{Hinton06,LeCunECCV2010,AndrewNgCVPR2011, ImagenetNIPS2012, DSPACV13, rbgCVPR2014}. To address human action recognition from videos, Ji et al.~\cite{3DCNNPAMI} developed a novel deep architecture of convolutional networks, where they extracted features from both spatial and temporal dimensions. Amer and Todorovic~\cite{SinisaSPNCVPR2012} applied Sum Product Networks (SPNs) to model human activities based on variable primitive actions. Our deep structured model can be viewed as an extension of these existing architectures, in which we make the network reconfigurable during learning and inference.

\section{Structured Deep Model}

In this section, we firstly introduce the structure of our deep structured model, and then explain how it can handle large intra-class variance with the latent structure.

\vspace{3mm}
\subsection{Spatio-temporal CNNs}

Our deep model is presented as a spatio-temporal convolutional neural network, as shown in Figure~\ref{fig:Architecture}. To model the complex human activities, it comprises of $M$ network cliques, which jointly conduct the final output. We define a clique as a subpart of the network stacked up for several layers. In particular, each clique extracts features from one decomposed video segment associated to one separated sub-action from the complete activity, and an illustration is highlighted in Figure~\ref{fig:Architecture}. Specifically, for each clique, two 3D convolutional layers are first built upon the raw input (i.e. grayscale and depth data), which consists with at most $m$ video frames, and then followed by one 2D convolutional layer. Note that a max-pooling operator is applied on each 3D convolutional layer making our model robust to local body deformations and surrounding noises.  Afterwards, the convolution results generated by different cliques are merged and concatenated into a long feature vector, upon which we build two full connection layers to associate with the activity labels.  In the following, we introduce the detailed definitions for these components of our model.

\begin{figure}[!ht]
\centering
\includegraphics[width=3.0in]{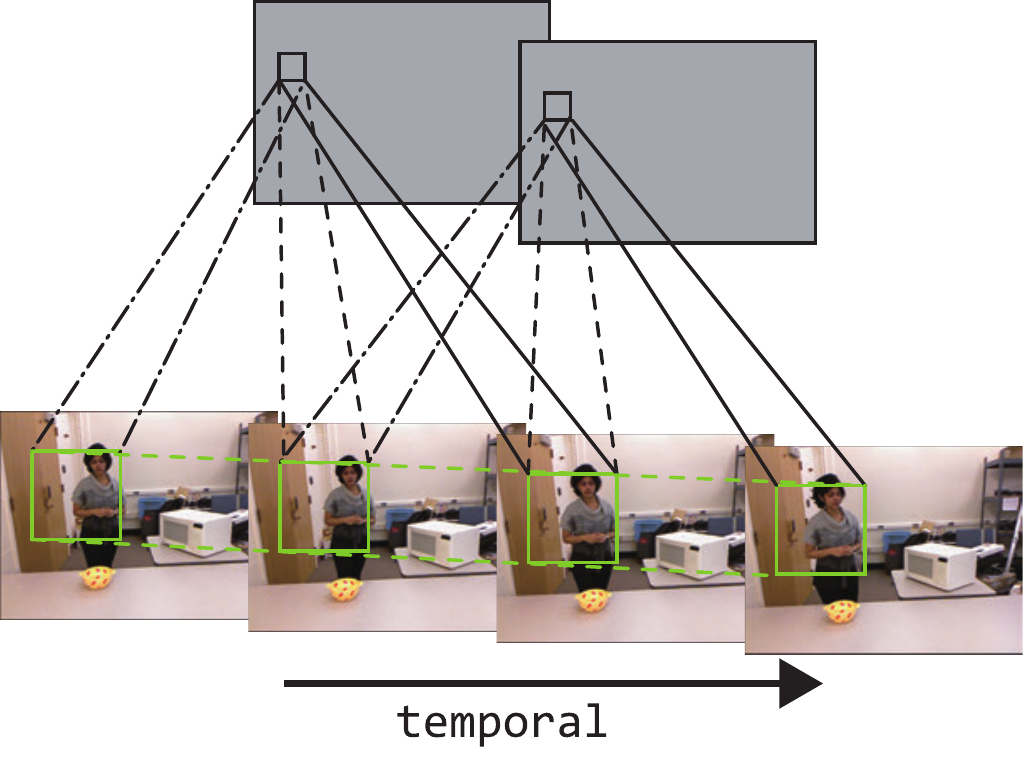}
\caption{Illustration of the 3D convolution across both spatial and temporal domains. In this example, the temporal dimension of the 3D kernel is 3. That is to say, each feature map is obtained by performing 3D convolutions across 3 adjacent frames.}\label{fig:conv3d}
\end{figure}

\textbf{3D Convolutional Layer.} The 3D convolutional kernels in one clique are computed independently to those from different cliques. For example, the kernels belonging to the first clique will only be applied to perform convolutions on the first temporal segment of the activity. Motivated by~\cite{3DCNNPAMI}, we perform the 3D convolutions spanning over both spatial and temporal dimensions of the input videos, and our model thereby captures both appearance and motion information for the observations. Suppose the width and the height of each frame are $w$ and $h$, and the size of the 3D convolutional kernel is $w^{\prime} \times h^{\prime} \times m^{\prime}$, where $w^{\prime}$, $h^{\prime}$, $m^{\prime}$ represents the width, height and temporal length, respectively. As Figure~\ref{fig:conv3d} illustrates, we can obtain a feature map via performing 3D convolutions across the $s$th to the $(s+m^{\prime}-1)$th frames. The response for the position $(x,y)$ in the feature map can be represented as,
\begin{eqnarray} \label{eq:3Dconv}
v_{xys} = \tanh(b + \sum_{i=0}^{w^{\prime}-1} \sum_{j=0}^{h^{\prime}-1} \sum_{k=0}^{m^{\prime}-1} \omega_{ijk} \cdot p_{(x+i)(y+j)(s+k)}),
\end{eqnarray}
where $p_{(x+i)(y+j)(s+k)}$ is the input pixel value at position $(x+i,y+j)$ in the $(s+k)$th frame, $\omega_{ijk}$ is the parameter for the convolutional kernel, and $b$ is the bias for the feature map. Thus we can obtain $m - m^{\prime} + 1$ feature maps, each with size of $(w - w^{\prime}+ 1, h - h^{\prime}+ 1)$. Note that one convolutional kernel only extracts one kind of feature. Thus we employ several kernels to generate different kinds of feature in each convolutional layer. For each model clique, we define that the number of 3D convolutional kernels in the first and second layers as $c_1$ and $c_2$.

After the first 3D convolutions, we obtain $c_1$ sets of $m - m^{\prime} + 1$ feature maps. For each set of feature maps, we further perform 3D convolutions on it, and generate another set of feature maps on a deeper layer. Note that as we employ $c_2$ kernels on the $c_1$ sets of feature maps, we can obtain $c_1 \times c_2$ sets of new feature maps in the next layer.

\begin{figure}[!htb]
\centering
\includegraphics[width=\columnwidth]{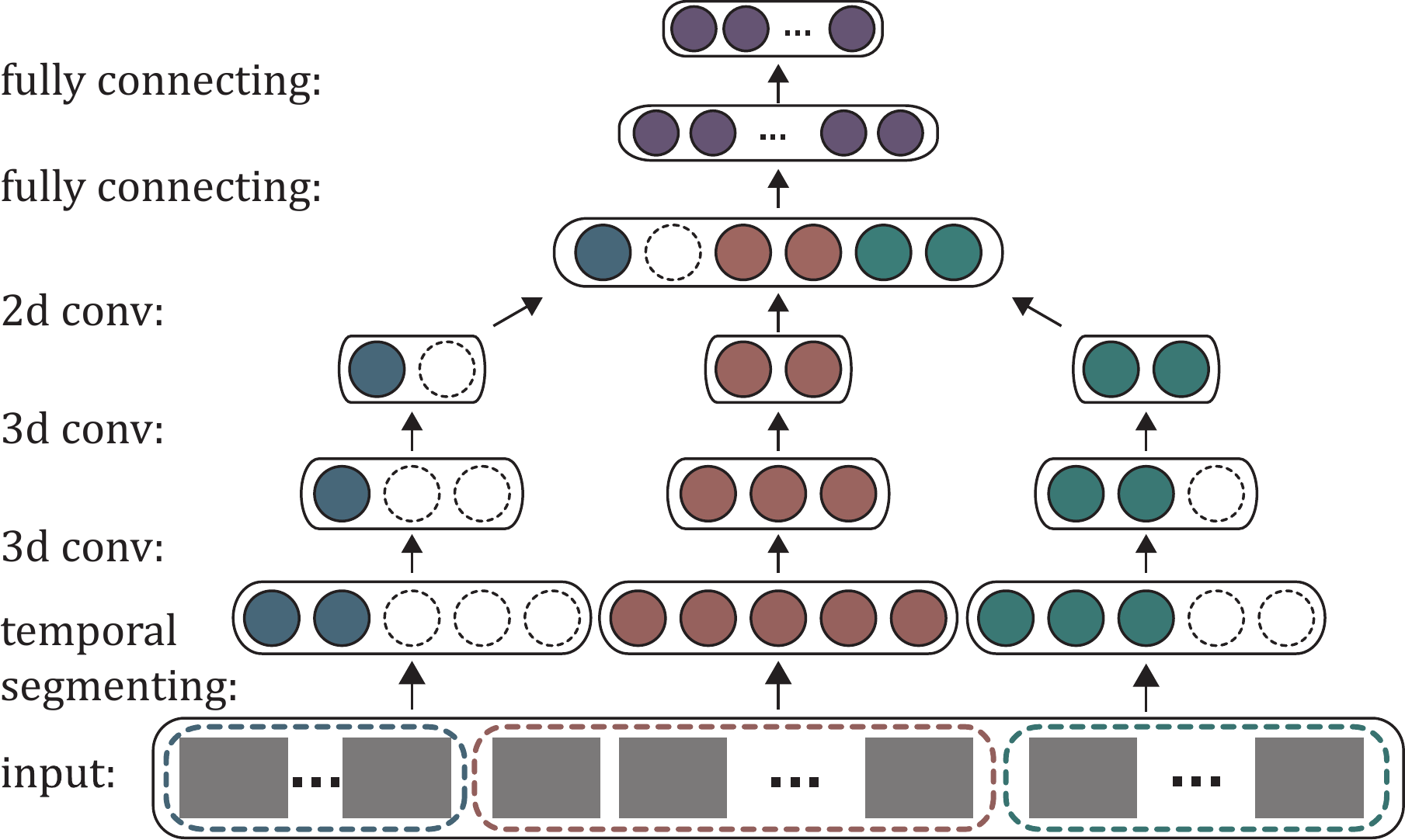}
\caption{ Illustration of our deep model incorporating the latent structure. Different model cliques are represented by different colors. As the temporal segmentation for an input video can be variant, different cliques might have different number of input frames. For the cliques whose input frame number less than $m$, part of the neurons in the network are inactivated, as represented by the dotted blank circles. }\label{fig:latent_structure}
\end{figure}

\textbf{Max-pooling Operator.} In our model, we apply a max-pooling operator after each 3D convolution result. This is a procedure widely applied to obtain deformation and shift invariance~\cite{ImagenetNIPS2012,KaiYuSC}. Given a set of feature maps, the max-pooling operator performs subsampling on them, which leads to the same number of feature maps with lower spatial resolution. More specifically, if a $2 \times 2$ max-pooling operator is performed on a $a_1 \times a_2$ feature map, we collect the max value in each $2 \times 2$ non-overlap regions to form a new feature map with size of $a_1 / 2 \times a_2 / 2$.

\textbf{2D Convolutional Layer.} After two layers of 3D convolution followed with max-pooling, each set of feature maps reduces to relatively a small temporal dimension. We then further apply 2D convolutional kernels to extract higher-level complex features from them. The 2D convolution can be viewed as a special case of 3D convolution by setting the temporal dimension of the 3D kernel to one, i.e. $m^{\prime}  = 1$. By performing 2D convolution on a set of feature maps, we can obtain the same number of feature maps in a new set. Suppose the number of 2D convolutional kernels is $c_3$, by performing 2D convolutions on $c1 \times c2 $ sets of feature maps, we can obtain $c_1 \times c_2 \times c_3$ sets of new feature maps.

\textbf{Full Connection Layer.} There are two full connection layers in our model. We first concatenate different sets of feature maps from the $M$ model cliques into a long feature vector. Then each unit of this vector are further connected with all neurons in the first full connection layer, which are further fully connected with output neurons. Note that the number of the output neurons is $K$, which is the same as the number of categories of activities $K$, and each of the neurons represents the probability of an activity hypothesis. To normalize the probabilities of the output labels, we apply the softmax function on them,
\begin{eqnarray} \label{eq:softmax}
\sigma(z_i) = \frac{\exp{(z_i)}}{\sum_{k=1}^K \exp{(z_k)}},
\end{eqnarray}
where $z_i$ is the $i$th value computed by multiplying the neuron values in the second last layer with the weights connected to the $i$th output neuron, and $\sigma(z_i)$ is the output probability. Note that $\sum_{i=1}^K \sigma(z_i) = 1$.

\emph{Input Data Details.} In our experiment, we obtain the gray and depth image from the raw data, which are taken from the gray channel and the depth channel of each video frame. To perform convolutions, we duplicate the channels for the 3D convolutional kernels in the first layer. The convolution results for these two channels are summed up together, thus the dimensions of the convolved feature maps remain the same. Note that our model can be generalized to apply on multi-channel video frames.

\subsection{Reconfigurable Latent Structure}

One key contribution of this paper is incorporating latent structure in deep model. Given different activity videos, the starting anchor frame and the number of input frames for each model clique can be variant. To illustrate it, we present a toy example in Figure~\ref{fig:latent_structure}, in which three model cliques represented by different colors are presented. Accordingly, the whole activity is decomposed into three action segments. The starting frame for each clique is flexible and the video is not evenly segmented. In this scheme, a clique is trained to handle the missing data by inactivating parts of the neurons. In other words, for the clique whose input anchor frame number is less than $m$, part of the neurons in the network are inactivated, as represented by the dotted blank circles in the first and third model cliques. Given input segments, classification on activities can be achieved with the deep model by performing forward propagation.

Formally, given a video sample, we define the index of starting anchor frames for $M$ cliques as $(s_1,...,s_M)$ and the corresponding number of input frames are $(t_1,...,t_M)$ where $1 \leq t_i \leq m$. Thus, the latent variables for our model can be represented as $H=(s_1,...,s_M, t_1,...,t_M)$, which infers the video segments for each model clique. Given the input video $X$, latent variables $H$, and model parameters $\omega$ (including different layer parameters and biases), we represent the classification results obtained by our model as $F(X,\omega,H)$, which is a vector of the probabilities for each activity. For simplicity, we define the probability for the $i$th activity as $F_i(X,\omega,H)$.

Note that our model explicitly decomposes the input activity into sub-actions by incorporating the latent variables. But the sub-actions will be not directly co-related with semantic meaning. The different values of $M$ (i.e. the number of sub-actions for one activity) will affect the final classification performance. 
When there is only one clique, the reconfigurability would be disregarded, and our model is simplified as the traditional CNN such as \cite{3DCNNPAMI}, which is compared in Sect.~\ref{sec:exper}. We can empirically tune $M$, just like setting the number of parts for the deformable part-based model in object detection. The model with a small $M$ could be less expressive to handle activity temporal variations, while a large $M$ could lead to over-fitting for the high model complexity. In practice, we can roughly estimate $M$ using traditional parameter tuning methods, e.g., the cross validation. We have shown the effectiveness of our method by setting $M$ = 4 on the databases.

\section{Learning}

As our deep model incorporates latent structure, the standard back propagation algorithm~\cite{CNN1990} is not applicable to optimize the model parameters. Thus we propose an EM-type optimization algorithm, namely Latent Structural Back Propagation (LSBP), to learn our model. Due to the large number of our model parameters and the insufficience of RGB-D data in human activities, we introduce a pre-training scheme to borrow the data strength from 2D data to optimize our model.

\begin{figure}[!htb]
\centering
\includegraphics[width=\columnwidth]{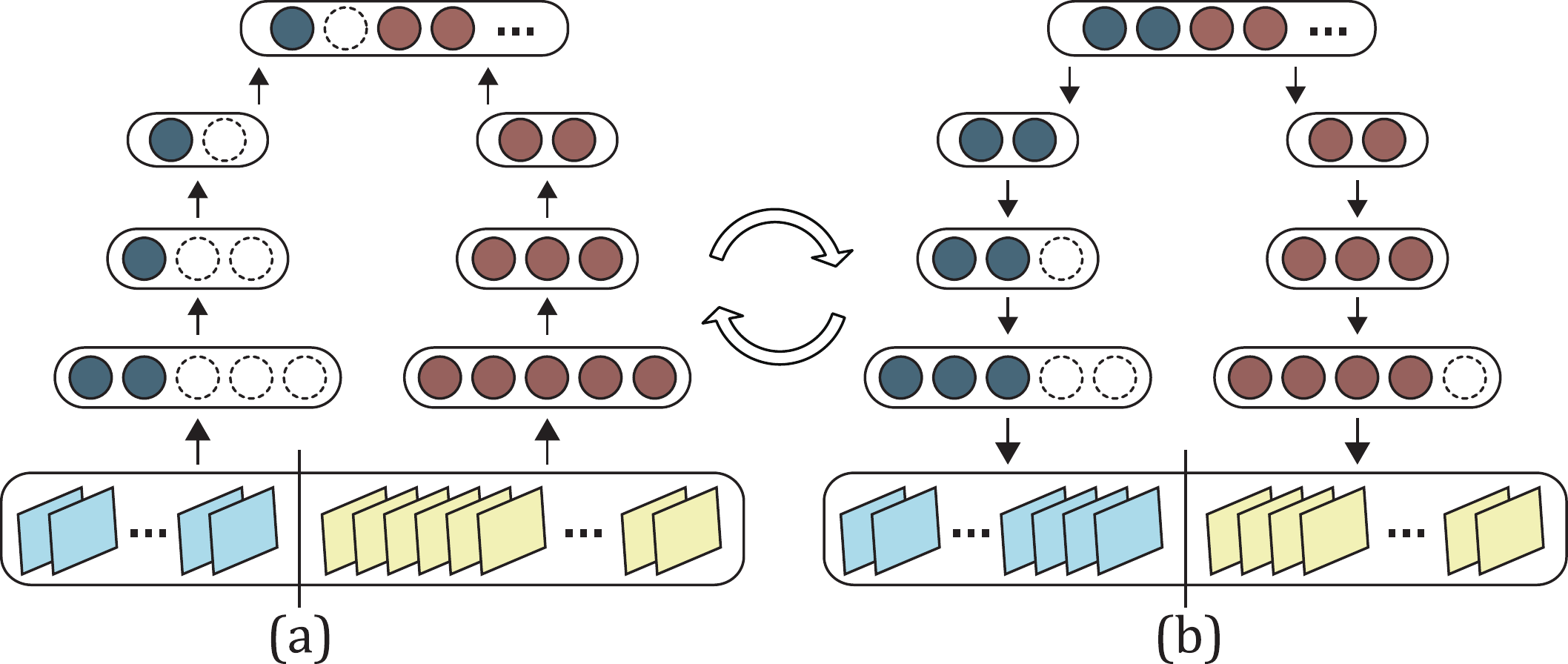}
\caption{ Illustration for the Latent Structural Back Propagation (LSBP). It is an iterative algorithm with two steps: (a) Given the current model parameters $\omega$, estimate the latent variables $H$ by adjusting the video segmentation; (b) Given the estimated latent variables $H^{*}$ which infers a video segmentation proposal, perform back propagation to optimize model parameters $\omega$. Note that the network neurons can be partially inactivated (as the dotted circles) according to the input segments.}\label{fig:learning}
\end{figure}
\subsection{Latent Structural Back Propagation}

Given a video with different latent variables, the input segments for each model clique will be different. During learning, the model parameters $\omega$ and the latent variables $H$ are required to be optimized at the same time. We propose an EM-type algorithm to iteratively optimize $\omega$ and $H$ in two steps: (i) Given the model parameters $\omega$ we can compute the latent variables $H$ (Figure~\ref{fig:learning}.(a)); (ii) Given the input frames decided by $H$, we can perform back propagation to optimize the model parameters $\omega$ (Figure~\ref{fig:learning}.(b)).

Suppose there are a set of $N$ training samples $(X_1,y_1)$, ... , $(X_N,y_N)$, where $X$ is the video, $y \in \{1,...,K\}$ represents the activity classes and $K$ is the number of activity classes. For simplicity, we also define the set of latent variables for all samples as ${\bf{H}} = \{H_1,...,H_N\}$. We apply the logistic regression to define the cost function $J(\omega,{\bf{H}})$ in training, which is defined as,
\begin{eqnarray} \label{eq:Cost}
J(\omega,{\bf{H}}) &=& - \frac{1}{N} (\sum_{i=1}^N \sum_{k=1}^K {\bf{1}}(y_i=k) \log F_k(X_i, \omega, H_i) \nonumber \\
&&+ (1 - {\bf{1}}(y_i=k)) \log(1 - F_k(X_i, \omega, H_i)) ) \nonumber \\
&& + ||\omega||^2,
\end{eqnarray}
where ${\bf{1}}(\cdot) \in \{0,1\}$ is the indicator function. The first two terms in Eq.(\ref{eq:Cost}) are the opposite of the log-likelihood and the last one is the regularization term. To minimize the cost $J(\omega,{\bf{H}})$, we optimize parameters $\omega$ and ${\bf{H}}$ in a 2-steps iteration as below.

{\textbf{(i)}} Given the model parameters $\omega$ obtained from the last iteration, we can minimize Eq.(\ref{eq:Cost}) by maximizing the probability $F_{y_i}(X_i, \omega, H_i)$ for each sample $(X_i, y_i)$, which is achieved by finding the most appropriate latent variable $H$,
\begin{eqnarray} \label{eq:EMH}
H_i^{*} = argmax_{H_i} F_{y_i}(X_i, \omega, H_i).
\end{eqnarray}
Recall that we apply softmax operator on the output results as Eq.(\ref{eq:softmax}), thus the maximization of $F_{y_i}(X_i, \omega, H_i)$ also depresses the probabilities of other labels $F_k(X_i, \omega, H_i), \forall k \neq y_i$. It leads to the increase of the log-likelihood and decrease the cost $J(\omega,{\bf{H}})$.

{\textbf{(ii)}} Given the latent variables  ${\bf{H}} = \{H_1,...,H_N\}$ for each sample, we can obtain the input sample segments for the deep structured model. After computing the cost $J(\omega,{\bf{H}})$ with the current inputs, we can obtain the gradients of $J(\omega,{\bf{H}})$ with respect to parameters $\omega$. By performing the back propagation algorithm, we can further decrease the cost $J(\omega,{\bf{H}})$ and optimize the model parameters $\omega$. Note that during back propagation, we apply stochastic gradient descent to update the parameters, and the update stops when it runs through all the training samples for one time.

The optimization algorithm iterates between these two steps until Eq.(\ref{eq:Cost}) converges.

\subsection{Model Pre-training}
To handle the insufficient RGB-D data for training and to boost the activity recognition results, we apply a pre-training scheme to initialize our model before optimizing the parameters with our LSBP algorithm on 3D data.

Given the large sum of 2D activity videos with labels, we train our deep model in a supervised manner. In this procedure, we first initialize the model parameters randomly, and each 2D video is evenly segmented to the number as the number of model cliques. To train on the 2D data, we directly apply back propagation instead of the proposed LSBP algorithm. It is mainly because of the following two reasons: (i) The initial model parameters are unstable, and it is not reliable to estimate latent variables $H$ with them; (ii) The training efficiency can be improved without considering the latent variables.

After training on the 2D data, we apply the parameters of the convolutional layers to initialize our model.  Note that we have gray and depth channels for each input frame in 3D data and only one gray channel for the 2D data. We thus duplicate the dimension of the 3D convolutional kernels in the first layer and initialize the parameters for the depth channel by the parameters for the gray channel. For the full connection layers, we set the parameters to random values. The reason is that we only need to borrow the feature learned in the 2D data, because the higher level information should be learned directly from the specific 3D activity dataset.

We summarize the whole learning procedure as Algorithm \ref{alg:Framwork}.

\begin{small}
\begin{algorithm}[htb]
\caption{Learning Framework}
\label{alg:Framwork}
\begin{algorithmic}\footnotesize
\REQUIRE ~~\\
    The labelled 2D and 3D activity dataset.
\ENSURE ~~\\
    Model parameters $\omega$.

\INPUT ~~\\
    Pre-train the Spatial-Temporal CNN in the 2D dataset.
    \\
\vspace{0.5em}
\hspace{-1.5em} Learning on 3D dataset:
\MYWHILE
    \STATE
    \begin{itemize}
\setlength{\itemsep}{1pt}
 \setlength{\parskip}{0pt}
 \setlength{\parsep}{10pt}
      \item[1.] Estimate the latent variables ${\bf{H}}$ by fixing model parameters $\omega$.
      \item[2.] Optimize $\omega$ given the input sample segments indicated by ${\bf{H}}$.
    \end{itemize}
\MYENDWHILE {$J(\omega,{\bf{H}})$ in Eq.(\ref{eq:Cost}) converges. }

\end{algorithmic}
\end{algorithm}
\end{small}

\section{Inference}
%

The inference task is to recognize the category of the activity given a video $X$. Formally, we perform the standard procedure of brute search for the activity label $y$ and the latent variables $H$ by maximizing the probability of $F_i(X,\omega,H)$,
\begin{eqnarray} \label{eq:inference}
(y^{*},H^{*}) = argmax_{(y,H)} F_y(X,\omega,H).
\end{eqnarray}
To do this, we search across all the labels $y(1 \leq y \leq K)$ and calculate the maximum probability $F_y(X,\omega,H)$ by optimizing $H$. Given the domain space of $H=(s_1,...,s_M, t_1,...,t_M)$, we constrain the input frame number for each model clique as $\tau \leq t_i \leq m$, and different video segments should not have overlaps (i.e., $s_i + t_i \leq s_{i+1}$). In all our experiments, we set the constant $\tau= 5$ during training and inference. We enumerate all the possibilities of $H$ under these constraints, and calculate the corresponding probabilities $F_y(X,\omega,H)$ via forward propagations. By selecting the highest probability, we obtain the optimal $F_y(X,\omega,H^{*})$. Though this optimization is a procedure of brute search, we can take advantage of parallel computation with GPU for that the forward propagations decided by different $H$ are independent of each other. In our experiment, parallel computation via GPU highly accelerates the inference speed.

\begin{figure*}[!htb]
\centering
\includegraphics[width=\textwidth]{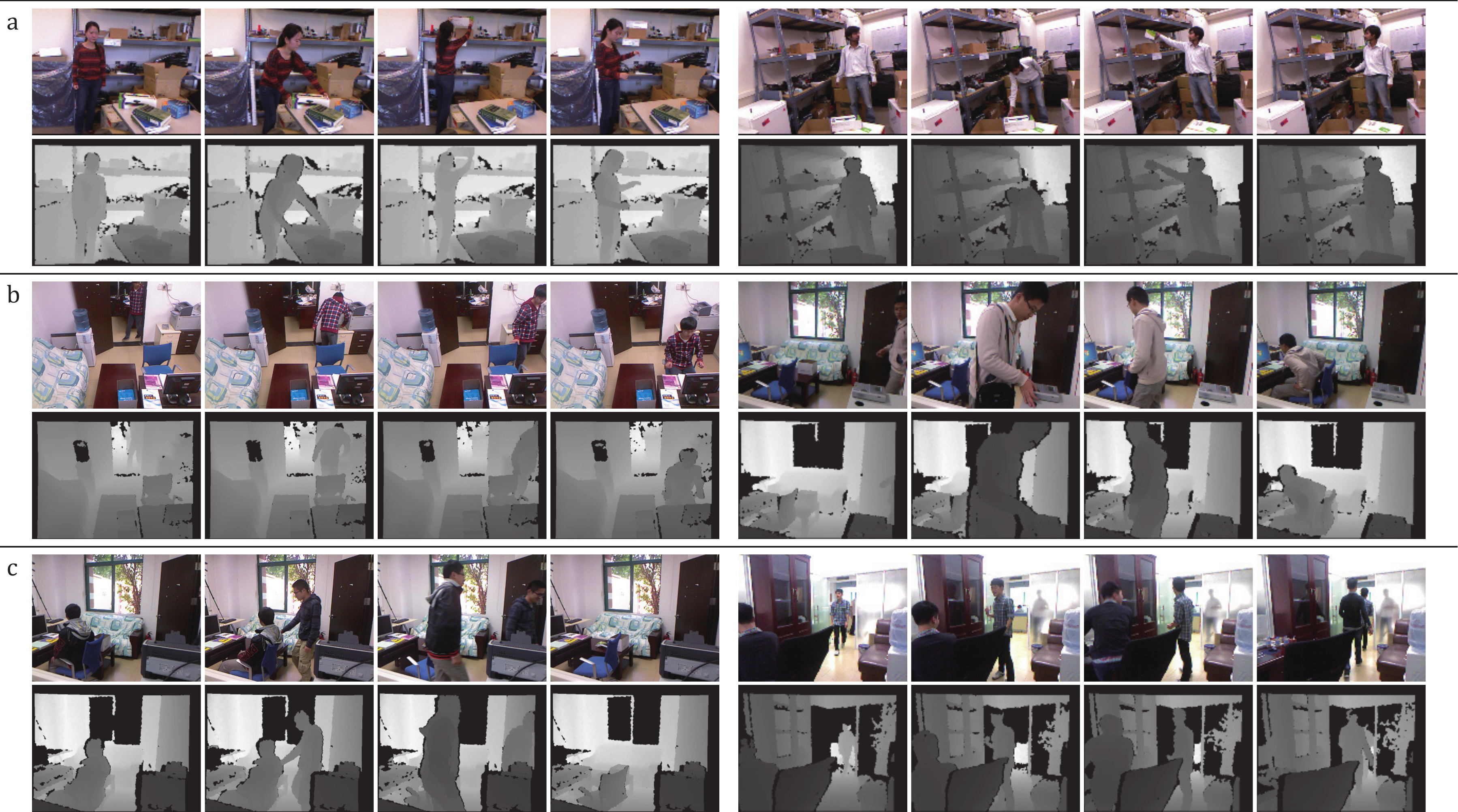}
\caption{Activity examples from the testing databases. Several sampled frames and depth maps are presented. (a) CAD-120, (b) OA1, (c) OA2, respectively, show two activities of the same category selected from the three databases.}\label{fig:dataset}
\end{figure*}

\section{Experiment}
\label{sec:exper}
We validate our approach on the CAD-120 activity dataset ~\cite{CADIJRR2013} and a larger dataset newly created by us, namely Office Activity (OA) dataset. Comparisons with state-of-arts and empirical analysis are presented in the experiments as well.

The CAD-120 dataset contains 120 RGB-D activity sequences of $10$ categories, which is widely used in 3D human activity recognition. These activities were performed by four different subjects, and each activity was repeated three times by the same actor. The challenges on this dataset lie in the large variance in object appearance, human pose, and viewpoint. The proposed OA dataset is more comprehensive, covering the regular daily activities taken place in an office, and it is the largest activity dataset of RGB-D videos, which includes $1180$ sequences. Three RGB-D sensors (i.e. Microsoft Kinect cameras) are utilized to capture data in three different viewpoints, and more than 10 actors are involved. The activities are captured in two different offices to increase the variability, where each actor performs the same activity twice for each viewpoint. It is worth mentioning that this dataset also includes the activities performed by two subjects with interactions. In particular,  it is divided into two sub-sets, each of which contains 10 categories of activities: OA1 (activities by a single subject) and OA2 (activities by two subjects). The categories in OA1 are: \{{\em answering-phones, arranging-files, eating, moving-objects, going-to-work, finding-objects, mopping, sleeping, taking-water, wandering}\}, and in OA2, we have \{{\em asking-and-away, called-away, carrying, chatting, delivering, eating-and-chatting, having-guest, showing, seeking-help, shaking-hands}\}. The OA database is publicly accessible\footnote{http://vision.sysu.edu.cn/projects/3d-activity/}. Several sampled frames and depth maps from the databases are exhibited in Figure~\ref{fig:dataset}. 
Following \cite{CADIJRR2013}, we apply four-fold cross validation procedure for testing in the CAD-120 dataset. That is, the model is trained on the activities of three subjects and tested on a new subject each time. The final outputs are averaged on the results of all four validations. For the OA dataset, we apply five-fold cross validation in the similar way.

\subsection{Implementation}

Given the RGB-D videos, we first normalize them into the same temporal length as preprocessing.  For each video, we extract $120$ frames by removing the ones with similar appearances in gray-scale. Then we further obtain $30$ anchor frames with a step size of $4$ in the $120$ frames. Suppose we index the $120$ frames from $1$ to $120$, then the selected frames are indexed by $1,5,9,...,120$. We apply these $30$ anchor frames as the inputs for our model.

We scale the input frame to $w = 80$ and $h = 60$ in our experiments. The number of decomposed video segments (i.e. actions) is $M=4$, and the length of the maximum number of input frames of each segment is $m=9$. Recall that the actual input frames can be less than $m$, as we introduce the latent variables manipulating the temporal decomposition. The networks in each clique are of the same structures (e.g. kernels). For each clique, the number of 3D convolutional kernels in the first layer is $c_1=7$, and the size of the kernel is $9 \times 7 \times 3$, where each number represents the width, height and temporal length. In the second layer, the number of 3D kernels is $c_2=5$, and the size is $7 \times 7 \times 3$. We apply $3 \times 3$ max-pooling operator over the 3D convolutions. In the 2D convolutional layer, we have $c_3=4$ kernels with size of $6 \times 4$. Hence we can obtain $700$ feature maps with size $1 \times 1$ as the outputs for each network clique, and we merge the feature maps  together into a vector of $700 \times 4 = 2800$ dimensions. Each unit in this vector is linked to $64$ neurons in the next full connection layer. At last, the $64$ neurons are fully connected to associate with the activity labels.

\begin{figure}[!htb]
\centering
\includegraphics[width=2.6in]{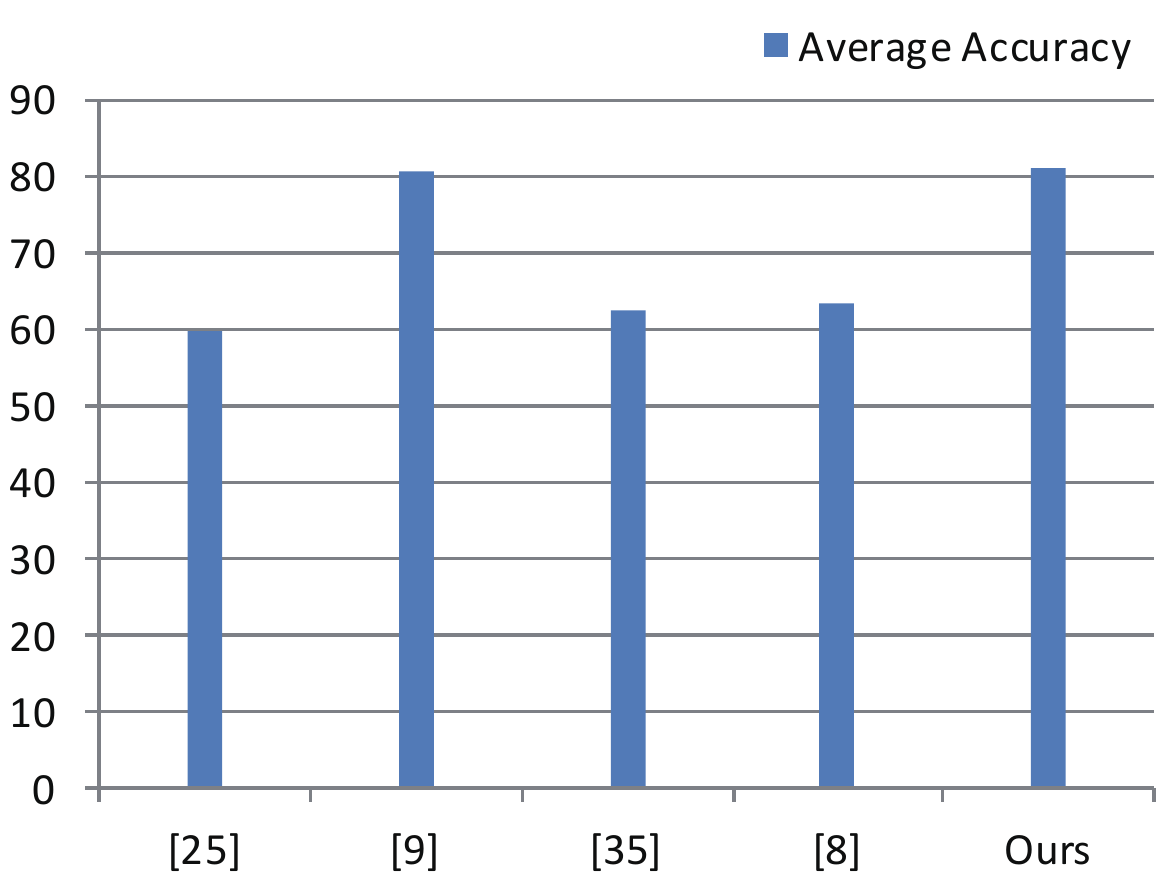}
\caption{The average accuracy on the CAD-120 database.}\label{fig:CAD_results}
\end{figure}

\begin{table}[!htbp]
\center
\begin{tabular}{|c|c|c|c|}
\hline
\hline
& Xia et al~\cite{DSTIP}. & Ji et al~\cite{3DCNNPAMI}. & Ours \\
\hline
\hline
arranging-objects & 75.0\% & 68.3\% & \textbf{82.3}\% \\
cleaning-objects & 68.3\%  & 60.0\% & \textbf{79.7}\% \\
having-meal & 41.7\%  & 60.0\% & \textbf{71.0}\% \\
making-cereal & 76.7\% & 77.6\% & \textbf{91.5}\% \\
microwaving-food &  36.7\% & 71.7\% & \textbf{85.3}\% \\
picking-objects & 75.0\% & 58.3\% & \textbf{97.2}\% \\
stacking-objects & \textbf{75.0}\% & 48.3\% & 61.0\% \\
taking-food & 83.3\% & 73.3\% & \textbf{93.5}\% \\
taking-medicine & 58.3\%  & 76.7\% & \textbf{96.8}\% \\
unstacking-objects & 33.3\% & 36.7\% & \textbf{54.0}\% \\
\hline
\hline
Average accuracy  & 62.3\% & 63.1\% & \textbf{81.2}\% \\
\hline
\end{tabular}
\caption{Accuracy of all categories on the CAD120 dataset. }
\label{tab:cad120}
\end{table}

\begin{table}[!hbp]
\center
\begin{tabular}{|c|c|c|c|}
\hline
\hline
& Xia et al~\cite{DSTIP} & Ji et al\cite{3DCNNPAMI} & Ours \\
\hline
\hline
answering-phones & 12.5\% & \textbf{40.0}\% & 35.0\% \\
arranging-files & 59.7\% & 53.3\% & \textbf{84.4}\% \\
eating & 40.3\% & 41.7\% & \textbf{65.5}\%\\
moving-objects & 48.6\% & 51.7\% & \textbf{61.1}\% \\
going-to-work & 34.7\% & 41..7\% & \textbf{92.2}\% \\
finding-objects & \textbf{65.3}\% & 36.7\% & 53.9\% \\
mopping & 63.9\% & 66.7\% & \textbf{72.2}\% \\
sleeping & 25.0\% & \textbf{45}\% & 43.9\% \\
taking-water & \textbf{58.3}\% & 40.0\% & 51.7\% \\
wandering & \textbf{56.9}\% & 50.0\% & 40.6\% \\
\hline
\hline
Accuracy & 46.5\% & 46.7\% & \textbf{60.1}\% \\
\hline
\end{tabular}
\caption{ Quantitative results on the OA1 dataset.  Accuracy per activity category and average accuracy of all categories are reported.}
\label{tab:hao1}
\end{table}

\begin{table}[!hbp]
\center
\begin{tabular}{|c|c|c|c|}
\hline
\hline
& Xia et al~\cite{DSTIP} &Ji et al\cite{3DCNNPAMI} & Ours \\
\hline
asking-and-away & 12.5\% & \textbf{39.6}\% & 25.3\% \\
called-away & 45.8\% & 44.8\% & \textbf{57.5}\% \\
carrying & \textbf{66.7}\% & 56.8\% & 53.5\% \\
chatting & 37.5\% & 17.2\% & \textbf{25.3}\% \\
delivering & 20.1\% & \textbf{34.5}\% & 32.8\% \\
eating-and-chatting & 50.0\% & 35.8\% & \textbf{69.5}\% \\
having-guest & 37.5\% & 34.1\% & \textbf{43.7}\% \\
seeking-help & 16.7\% & 44.8\% & \textbf{59.2}\% \\
shaking-hands & 41.7\% & 32..8\% & \textbf{59.8}\% \\
showing & \textbf{37.5}\% & 29.3\% & 23.0\% \\
\hline
\hline
Accuracy & 36.6\% & 37.0\% & \textbf{45.0}\% \\
\hline
\end{tabular}
\caption{ Quantitative results on the OA2 dataset. Accuracy per activity category and average accuracy of all categories are reported.}
\label{tab:hao2}
\end{table}

\begin{table}[!hbp]
\center
\begin{tabular}{|c|c|c|c|}
\hline
  & grayscale & depth & grayscale + depth \\
\hline
OA1 & 44.9\% & 57.2\% & 60.1\% \\
OA2 & 41.6\% & 43.6\% & 45.0\% \\  
\hline
\end{tabular}
\caption{ Channel analysis on the two dataset. Average accuracy of all categories are reported.}
\label{tab:channel}
\end{table}

The experiments are executed on a desktop with an Intel i7 3.7GHz CPU, 8GB RAM and GTX TITAN GPU. For model learning, we set the learning rate as $0.002$ for applying the stochastic gradient descent algorithm. The training times of one fold are 2 hours for CAD-120 (including 120 videos and occupying 0.2GB), 5 hours for OA1 (including 600 videos and occupying 1.0GB), and 5 hours for OA2 (including 580 videos and occupying 0.9GB) dataset, respectively. Each iteration of training costs similar time, and the convergence of our model over iterations is shown in Figure.~\ref{fig:pretrain} and~\ref{fig:latentStruct}. For inference, as we fully take advantage of parallel computation with GPU, it only takes around $0.4$ seconds to complete recognition on a given video.


\subsection{Results and Comparisons}

\emph{CAD-120 dataset.} On this dataset, we adopt four state-of-the-art methods for comparison. As shown in Figure~\ref{fig:CAD_results}, our approach obtains the average accuracy of $81.2\%$, distinctly superior than results generated by other four competing methods, such as $59.7\%$~\cite{SungICRA2012}, $80.6\%$~\cite{CADIJRR2013}, $62.3\%$~\cite{DSTIP} and $63.1\%$~\cite{3DCNNPAMI}. In Table~\ref{tab:cad120}, we report the detailed accuracies on all the categories, compared with the method based on hand-crafted feature engineering~\cite{DSTIP}, and the deep architecture of convolutional neural networks~\cite{3DCNNPAMI}. Note that for different methods we train the models using the same data annotation, which only includes the activity labels on videos.

\emph{OA dataset.} In this experiment, we apply our approach on the two sub-sets, respectively. Our deep structured model outperforms the state-of-the-art methods on average. On the OA1 set, our approach outperforms on 5 out of 10 categories and obtains the highest average accuracy of $60.1\%$, as Table.~\ref{tab:hao1} reports. On the OA2 set, the average accuracy of our method is $45.0\%$ and we have 6 classes of activities achieve better results than the other two competing methods, as Table~\ref{tab:hao2} reports. By reviewing the results, we find the failure cases probably caused by the lack of contextualized scene understanding. For example, understanding the activities of {\em taking-water} and {\em sleeping} actually requires to extra higher level information. We will consider it in future work. Moreover, the depth data is very useful. Since the testing is performed on the new subject, we observe that large appearance variances existed in grayscale data lead to worse performance. The depth data has much smaller variance, and does help to capture the motion information. Table~\ref{tab:channel} illustrates that depth data can boost the performance a lot, especially in OA1.

\vspace{5mm}
\subsection{Empirical Analysis}

For further evaluation, we conduct two following empirical analysis under different settings.

(I) To clarify significance of using the pre-training, we discard the parameters trained on 2D videos and learn the model directly on the RGB-D data. Then we compare the model with the original version by the test error rate, which is defined as one minus the classification accuracy. This testing is implemented on the OA1 dataset. In Figure~\ref{fig:pretrain}, we visualize the test error rates with the increasing of iteration numbers during training.  Each test error rate on a specific iteration number is calculated by applying the currently trained model. It is shown that the model using the pre-training converges after $25$ iterations, while the other one without the pre-training requires $140$ iterations. More importantly, the pre-training can effectively reduce the error rate in the testing, say $8\%$ less than without the pre-training.

(II) We demonstrate the effectiveness of incorporating reconfigurable structure in the deep architecture. That is, we can fix the latent variables in our model and train it using the standard back propagation algorithm. In this model without latent structure, each network clique receives $m$ input frames that are evenly segmented from the video. In Figure~\ref{fig:latentStruct}, we visualize the test error rates with different iterations of the two models: structured v.s. non-structured, from the same initialization. We observe that the error rates of the structured model decrease faster and reach to the lower result, compared with the non-structured model. This experiment is executed on the OA1 dataset.

\section{Conclusions}

This paper studies a novel deep structured model by incorporating model reconfigurability into layered convolutional neural networks. This model has been shown to handle well realistic challenges in 3D activity recognition, and it enables us to perform recognition from raw RGB-D data rather than relying on hand-crafted features. Moreover, we consider two aspects in future work. First, we can integrate high level semantic information into our model to deal with more complicated events with underlying intentions. Second, we plan to deploy our model into cloud computing platforms so that thin clients can expediently access the ability of 3D human activity understanding.
%
%

\end{document}